\documentclass{article}

\usepackage{PRIMEarxiv}

\usepackage[utf8]{inputenc} 
\usepackage[T1]{fontenc}    
\usepackage{hyperref}       
\usepackage{url}            
\usepackage{booktabs}       
\usepackage{amsfonts}       
\usepackage{nicefrac}       
\usepackage{microtype}      
\usepackage{lipsum}
\usepackage{fancyhdr}       
\usepackage{graphicx}       
\graphicspath{{media/}}     
\usepackage{graphicx}
\usepackage{amssymb}
\usepackage{amsmath}
\usepackage{amsthm}
\usepackage{amsmath}
\usepackage{xcolor}
\usepackage{caption}
\usepackage{subcaption}
\usepackage{mathrsfs}
\usepackage[noend,ruled,vlined]{algorithm2e}

\title{NSGA-PINN: A Multi-Objective Optimization Method for  Physics-Informed Neural Network Training}

\author{ Binghang~Lu \\
	Department of Computer Science\\
	Purdue University\\
	305 N University St, West Lafayette, IN 47907, USA \\
	\texttt{lu895@purdue.edu} \\
	\And
	Christian~B.~Moya \\
	Department of Mathematics\\
	Purdue University\\
	150 N. University St, West Lafayette, IN, 47907, USA \\
	\texttt{cmoyacal@purdue.edu} \\
	\AND
	Guang~Lin \\
	Department of Mathematics and School of Mechanical Engineering \\
        Purdue University\\
	150 N. University St, West Lafayette, IN, 47907, USA \\
	\texttt{guanglin@purdue.edu} \\
}

\begin{document}
\maketitle

\begin{abstract}
	This paper presents NSGA-PINN, a multi-objective optimization framework for effective training of Physics-Informed Neural Networks (PINNs). The proposed framework uses the Non-dominated Sorting Genetic Algorithm (NSGA-II) to enable traditional stochastic gradient optimization algorithms (e.g., ADAM) to escape local minima effectively. Additionally, the NSGA-II algorithm enables satisfying the initial and boundary conditions encoded into the loss function during physics-informed training precisely. We demonstrate the effectiveness of our framework by applying NSGA-PINN to several ordinary and partial differential equation problems. In particular, we show that the proposed framework can handle challenging inverse problems with noisy data.
\end{abstract}


\section{Introduction}
\label{sec:introduction}

Physics-informed neural networks (PINNs), as proposed in the seminal paper by Raissi et al. in \cite{raissi2019physics}, have recently gained significant attention in the scientific machine-learning community. This is due to their ability to accelerate the simulation and discovery of complex dynamical systems. PINNs infer the unknown solution (i.e., the physics of interest) by incorporating the underlying governing equations of the system into the training loss function ~\cite{chen2021physics}. As a result, they have enabled solving various problems modeled by ordinary and partial differential equations (PDEs)~\cite{raissi2020hidden} ~\cite{ karniadakis2021physics}, which can be challenging for standard numerical approaches. Furthermore, the physics-informed framework has successfully tackled challenging inverse problems~\cite{mao2020physics} ~\cite{fernandez2018towards} by combining PINNs with data (i.e., scattered measurements of the states).

PINNs (Physics-Informed Neural Networks) incorporate multiple loss functions, including residual loss, initial loss, boundary loss, and, if needed, data loss for inverse problems. The most common approach to train PINNs is by optimizing the total PINN loss function using standard stochastic gradient descent (SGD) ~\cite{ruder2016overview} ~\cite{cheridito2021non} ~\cite{bottou1991stochastic} methods such as Adam, as discussed in ~\ref{Stochastic Gradient Decent}. However, with SGD methods, optimizing highly non-convex loss functions ~\cite{jain2017non} ~\cite{szu1986non} for neural network training can be challenging since there is a risk of getting trapped in numerous sub-optimal local minima, particularly when solving inverse problems or dealing with noisy data ~\cite{krishnapriyan2021characterizing}. Additionally, SGD only allows for satisfying initial and boundary conditions as soft constraints, which may limit the use of PINNs in the design, optimization, and control of complex systems.

Inspired by multi-objective optimization algorithms~\cite{konak2006multi} ~\cite{gunantara2018review} ~\cite{deb2016multi}, we approach the above problems by considering the training of PINNs as a multi-objective optimization problem. We propose the NSGA-PINN framework to escape local minima and satisfy the system’s constraints, such as initial and boundary conditions, as hard constraints ~\cite{lu2021physics}. Specifically, we use the Non-dominated Sorting Genetic Algorithm-II (NSGA-II)~\cite{996017} to exactly satisfy the losses and help the SGD methods escape local minima.

Our experimental results on inverse ordinary differential equation (ODE) and partial differential equation (PDE) problems show that the NSGA-PINN algorithm is effective in converging to an optimal solution with excellent generalization capabilities and is robust to noise.



The rest of the paper is organized as follows. First, in Section~\ref{sec:background-description}, we provide a brief introduction to the following background information: PINN, SGD method, and NSGA-II algorithm. Then, in Section~\ref{sec:proposed-method}, we describe our proposed NSGA-PINN method. In Section~\ref{sec:numerical-experiments}, we present experimental results using inverse ODE problem and PDE problems to study the behavior of NSGA-PINN. We also test the robustness of our method in the presence of noisy data. Our results are discussed in Section~\ref{sec:discussion}. Finally, we conclude the paper in Section~\ref{sec:conclusion}.
\section{Background}
\label{sec:background-description}
This section introduces the physics-informed neural network (PINN) framework, the stochastic gradient descent (SGD) method, and the non-dominated sorting genetic algorithm (NSGA-II) algorithm.
\subsection{Physics-Informed Neural Networks}
Consider computing data-driven solutions to partial differential equations (PDEs) of the general form:
\begin{align}
u_{t} + \mathscr{N} [u:\lambda]=0, \quad x \in \Omega, t \in [0,T]
\label{PINN equation}
\end{align}
Here, $u$ represents the solution of the PDE, $\Omega \subset \mathbb{R} ^{d}$ represents the spatial domain, and $\mathscr{N}[:]$ denotes a differential operator.

The goal of PINN is to learn a parametric surrogate $u_\theta$ with trainable parameters $\theta$ that approximates the solution $u$. To achieve this goal, a neural network is constructed and the total loss function of PINN is minimized. The total loss function consists of several components: residual loss, initial loss, boundary loss, and data loss, i.e,
\begin{align} \label{eq:PINN loss}
    \mathcal{L}_{total} = w_f \mathcal{L}_{res} + w_g \mathcal{L}_{ics} + w_j \mathcal{L}_{bc} + w_h \mathcal{L}_{data} .
\end{align}
We use the coefficients $w$ to balance the loss terms. Each loss term is calculated by applying the L2 approximation ~\cite{de1993structured} ~\cite{hure2019some} ~\cite{germain2022approximation}. In particular, $\mathcal{L}_{res}$ denotes the residual loss, which is the difference between the exact value of the PDE and the predicted value from the PINN deep neural network (DNN) ~\cite{zhang2016dnn} ~\cite{jagtap2020adaptive}:
\begin{align*}
\mathcal{L}_{res} = \frac{1}{N_r}\sum_{i=1}^{N_r}\lvert\lvert u_\theta(x_{i}^r) - u(x_{i}^r) \rvert \rvert ^2
\end{align*}
In the above, we only considered the problem in the spatial domain for the sake of simplicity. Extending it to the temporal domain is straightforward. Moreover, $u_\theta(x_{i}^r)$ represents the output value of the PINN DNN on a set of $N_r$ points sampled within the spatial domain $\Omega$. This can be computed using automatic differentiation methods~\cite{baydin2018automatic}. On the other hand, $u(x_{i}^r)$ denotes the true solution of the PDE.

The initial or boundary loss represents the difference between the true solution and the predicted value from the PINN DNN at the initial or boundary condition. For instance, the boundary loss (for a given boundary condition~$h$) at a set of $N_b$ boundary points $x_i^b$ is defined as follows:
\begin{align*}
 \mathcal{L}_{bc} = \frac{1}{N_b}\sum_{i=1}^{N_b} \lvert\lvert u_\theta(x_{i}^b) - h(x_{i}^b) \rvert \rvert ^2
\end{align*}

Furthermore, if we tackle inverse problems and have a set of $N_d$ experimental data points $y_i^d$, we can calculate the data loss using the PDE equation as follows:
\begin{align*}
 \mathcal{L}_{data} = \frac{1}{N_d}\sum_{i=1}^{N_d} \lvert\lvert u_\theta(x_{i}^d) - y_{i}^d \rvert \rvert ^2
\end{align*}

As shown in Equation~\ref{eq:PINN loss}, the loss value of a physics-informed neural network (PINN) is calculated as a simple linear combination with soft constraints. To the authors' knowledge, no optimization methods are available to treat the PINN loss function as a multi-objective optimization problem.

\subsection{Stochastic Gradient Decent}
\label{Stochastic Gradient Decent}
Stochastic Gradient Descent (SGD) and its variants, such as Adam, are the most commonly used optimization methods for training neural networks (NN). SGD uses mini-batches of data, which are subsets of data points randomly selected from the training dataset. This injects noise into the updates during neural network training, enabling exploration of the non-convex loss landscape. The optimization problem for SGD can be written as follows:
\begin{align} \label{eq:ADAM equation}
    \theta^* = \text{arg}\min_\theta \mathcal{L}(\theta;\mathcal{T}).
\end{align}
This paper focuses on a variant of Stochastic Gradient Descent (SGD) known as the Adaptive Moment Estimates (Adam) optimizer. Adam is the most popular and fastest method used in deep learning. The optimizer requires only first-order gradients and has very little memory requirement. It results in effective neural network (NN) training and generalization.

However, applying SGD methods to Physics-Informed Neural Network (PINN) training presents inevitable challenges. For complex non-convex PINN loss functions, SGD methods can get stuck in a local minimum, particularly when solving inverse problems with PINNs or dealing with noisy data.

\subsection{NSGA-II algorithm}
The Non-dominated Sorted Genetic Algorithm NSGA-II algorithm, proposed by Deb, K. et al. in~\cite{996017}, is a fast and elitist algorithm for solving multi-objective optimization problems. Like other evolutionary algorithms (EAs) ~\cite{yu2010introduction} ~\cite{van1998multiobjective}, NSGA-II mainly consists of a parent population and genetic operators such as crossover, mutation, and selection. To solve multi-objective problems, the NSGA-II algorithm uses non-dominated sorting to assign a front value to each solution and calculates the density of each solution in the population using crowding distance. It then uses crowded binary selection to choose the best solutions based on front value and density value. We use these functions in our NSGA-PINN method, and we will explain them in detail in Section~\ref{sec:proposed-method}.

\section{The NSGA-PINN Framework}
\label{sec:proposed-method}
This section describes the proposed NSGA-PINN framework for multi-objective optimization-based training of a Physics-Informed Neural Network (PINN).
\subsection{Non-dominated Sorting}
\SetKwComment{Comment}{/ }{ /}
\begin{algorithm}
\caption{Non-dominated sorting}

\textbf{inputs:} P\;
\For {$p \in P$} {
 n =  [] \tcp*{set of points be dominated by other points}
 S = [] \tcp*{set of points dominate other points}
\For{$q \in P$}{
\If{p $<$ q} {   
$S_p = S_p \cup \{q\}$  \tcp*{ q is dominated by p}
}
\uElseIf{q $<$ p}
{$n_p = n_p +1$}
}
\If{$n_p = 0$} {   
$p_{rank} = 1$\;
$F_1 = F_1 \cup \{p\}$  \tcp*{ assign p to the first front}
}
}
i = 1\;
\While{$F_i \neq \emptyset$ }{
$ Q = \emptyset$      \tcp*{ used to store the members of the next front}
\For {$p \in F_i$}{
\For {$q \in S_p$} {
$n_q = n_q - 1$\;
\If{$n_q = 0$} {
$q_{rank} = i + 1$\;
$Q = Q \cup \{q\}$   \tcp*{q is in the next front}
}
}
}
i =i+1\;
$F_i = Q$\;
}
\label{ADAM-NSGA algorithm}
\end{algorithm}
The proposed NSGA-PINN utilizes non-dominated sorting (see Algorithm~\ref{ADAM-NSGA algorithm} for more detailed information) during PINN training. The input $P$ can consist of multiple objective functions, or loss functions, depending on the problem setting. For a simple ODE problem, these objective functions may include a residual loss function, an initial loss function, and a data loss function (if experimental data is available and we are tackling an inverse problem). Similarly, for a PDE problem, the objective functions may include a residual loss function, a boundary loss function, and a data loss function.

In the EAs, the solutions refer to the elements in the parent population. We randomly choose two solutions in the parent population $p$ and $q$, if $p$ has a lower loss value than $q$ in all the objective functions, we define $p$ as dominating $q$. If $p$ has at least one loss value lower than q, and all others are equal, the previous definition also applies. For each $p$ element in the parent population, we calculate two entities: 1) domination count $n_p$, which represents the number of solutions that dominate solution $p$, and 2) $S_p$, the set of solutions that solution $p$ dominates. Solutions with a domination count of $n_p = 0$ are considered to be in the first front. We then look at $S_p$ and, for each solution in it, decrease their domination count by 1. The solutions with a domination count of 0 are considered to be in the second front. By performing the non-dominated sorting algorithm, we obtain the front value for each solution~\cite{996017}.

\subsection{Crowding-Distance Calculation}
In addition to achieving convergence to the Pareto-optimal set for multi-objective optimization problems, it is important for an evolutionary algorithm (EA) to maintain a diverse range of solutions within the obtained set. We implement the Crowding-distance calculation method to estimate the density of each solution in the population. To do this, first, sort the population according to each objective function value in ascending order. Then, for each objective function, assign infinite distance values to the boundary solutions, and assign all other intermediate solutions a distance equal to the absolute normalized difference in function values between two adjacent solutions. The overall crowding-distance value is calculated as the sum of individual distance values corresponding to each objective. A higher density value represents a solution that is far away from other solutions in the population.
    
\subsection{Crowded Binary Tournament Selection}
\SetKwComment{Comment}{/ }{ /}
\begin{algorithm}
\caption{Crowded binary tournament selection}

\textbf{inputs:} N\;
mating pool = []\;
divide solutions into the N/2 array each array has 2 elements\;
\While{sizeof (mating pool) $\neq$ N}{
\For{ele in arr}{
\If{$F_{ele[0]} < F_{ele[1]}$}   {  
mating pool $\gets F_{ele[0]}$  \tcp*{ select element with lower front value}
}
\uElseIf{$F_{ele[0]} = F_{ele[1]}$}{ 
\If{$D_{ele[0]} > D_{ele[1]}$}{
mating pool $\gets F_{ele[0]}$    \tcp*{ select element with higher density value}
}\uElseIf{$D_{ele[0]} < D_{ele[1]}$}{mating pool $\gets F_{ele[1]}$}\Else{mating pool $\gets random(ele[0],ele[1])$}

}
\Else{mating pool $\gets random(ele[0],ele[1])$} 
}
}
\label{Crowded binary tournament selection algorithm}
\end{algorithm}
The Crowded binary tournament selection, explained in more detail in Algorithm~\ref{Crowded binary tournament selection algorithm}, was used to select the best physics-informed neural network (PINN) models for the mating pool and further operations. Before implementing this selection method, we labeled each PINN model so that we could track the one with the lower loss value. The population of size $n$ was then randomly divided into $n/2$ groups, each containing two elements. For each group, we compared the two elements based on their front and density values. We preferred the element with a lower front value and a higher density value. In Algorithm~\ref{Crowded binary tournament selection algorithm}, $F$ denotes the front value and $D$ denotes the density value.

\subsection{NSGA-PINN Main Loop}
\SetKwComment{Comment}{/* }{ */}
\begin{algorithm}
\caption{Training PINN by NSGA-PINN method}
\textbf{Hyper-parameters:} parent population N and max generation number $\alpha$\;
$count = 0$ \;
Initialize the parent set $S$: the parent set has a number of N neural networks.\;
\While{$count < \alpha$}{

$f1 \gets res(nn)$ for nn in S\;
$f2 \gets ics(nn)$ for nn in S\;
$f3 \gets data(nn)$ for nn in S\;
$R_t = P_t \cup Q_t$     \;
$F_i$ = non dominated sorting(f1,f2,f3)      \;
crowding distance sorting$(F_i)$\;
mating pool $\gets$ crowded binary selection($F_i,\prec_{n}$)\;
$Q_t \gets$ ADAM optimizer (mating pool(i))\;

$count = count +1$\;
}
\label{ADAM-NSGA algorithm}
\end{algorithm}
The main loop of the proposed NSGA-PINN method is described in Algorithm~\ref{ADAM-NSGA algorithm}. The algorithm first initializes the number of PINNs to be used ($N$) and sets the maximum number of generations ($\alpha$) to terminate the algorithm. Then, the PINN pool is created with $N$ PINNs. For each loss function in a PINN, $N$ loss values are obtained from the network pool. When there are three loss functions in a PINN, $3N$ loss values are used as the parent population. The population is sorted based on non-domination, and each solution is assigned a fitness (or rank) equal to its non-domination level~\cite{996017}. The density of each solution is estimated using crowding-distance sorting. Then, by performing a crowded binary tournament selection, PINNs with lower front values and higher density values are selected to be put into the mating pool. In the mating pool, the ADAM optimizer is used to further reduce the loss value. The NSGA-II algorithm selects the PINN with the lowest loss value as the starting point for the ADAM optimizer. By repeating this process many times, the proposed method helps the ADAM optimizer escape local minima.

\section{Numerical Experiments}
\label{sec:numerical-experiments}
This section evaluates the performance of physics-informed neural networks (PINN) trained with the proposed NSGA-PINN algorithm. We tested our framework on both ordinary differential equation (ODE) and partial differential equation (PDE) problems. Our proposed method is implemented using the PyTorch library. For each problem, we compared the loss values of each component of the PINN trained with the NSGA-PINN algorithm to the loss values obtained from the PINN trained with the ADAM method, using the same neural network structure and hyperparameters. To test the robustness of the proposed NSGA-PINN algorithm, we added noise to the experimental data used in each inverse problem.

\subsection{Inverse pendulum problem} 
The algorithm was first used to train PINN on the inverse pendulum problem without noise. The pendulum dynamics are described by the following initial value problem (IVP):
 \begin{align}
   &\dot{\theta}(t) = \omega(t) \\
   &\dot{\omega}(t) = - k \sin \theta(t) \nonumber
\end{align}
where the initial condition is sampled as follows $(\theta(0), \omega(0)) = (\theta_0, \omega_0) \in [-\pi, \pi] \times [0, \pi]$ and the true parameter unknown parameter~$k= 1.0$. 

Our goal is to approximate the mapping using a surrogate physics-informed neural network: $\theta_0, \omega_0, t \mapsto \theta(t), \omega(t)$. For this example, we used a neural network for PINN consisting of 3 hidden layers and 100 neurons in each layer. The PINN training loss for the neural network is defined as follows:
\begin{align*}
   \mathcal{L} = \mathcal{L}_{res} + \mathcal{L}_{ics} + \mathcal{L}_{data}.
\end{align*}
To determine the total loss in this problem, we add the residual loss, initial loss, and data loss. We calculate the data loss using the mesh data $t_s$, which ranges from 0 to 1 (seconds) with a step size of 0.01. We fit this data onto the ODE to determine the data loss value accurately. For this problem, we set the parent population to 20 and the maximum number of generations to 20 in our NSGA-PINN.

\begin{figure}[htb!]
     \centering
        \includegraphics[width=.32\textwidth, height=3.cm]{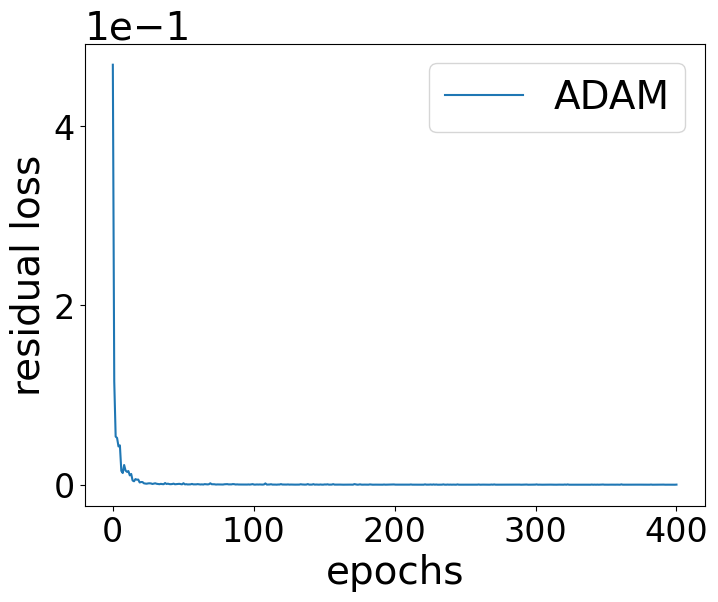} 
        \includegraphics[width=.32\textwidth, height=3.cm]{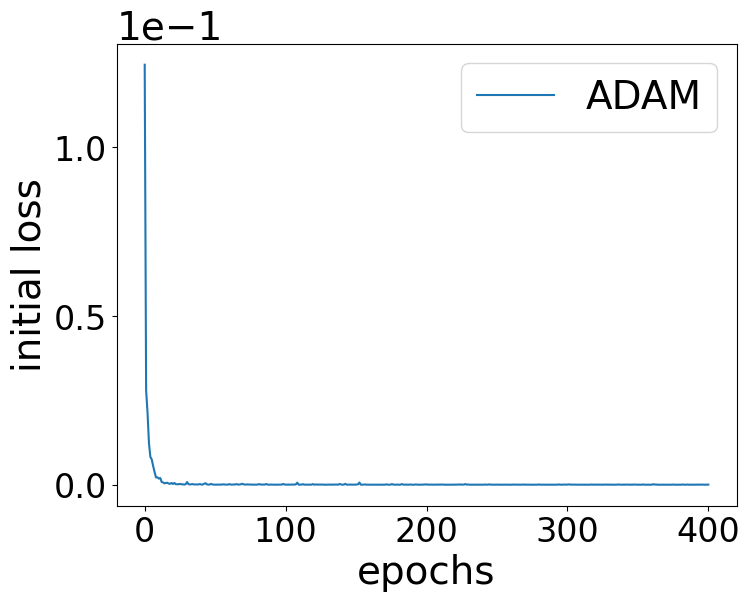}  
        \includegraphics[width=.32\textwidth, height=3.cm]{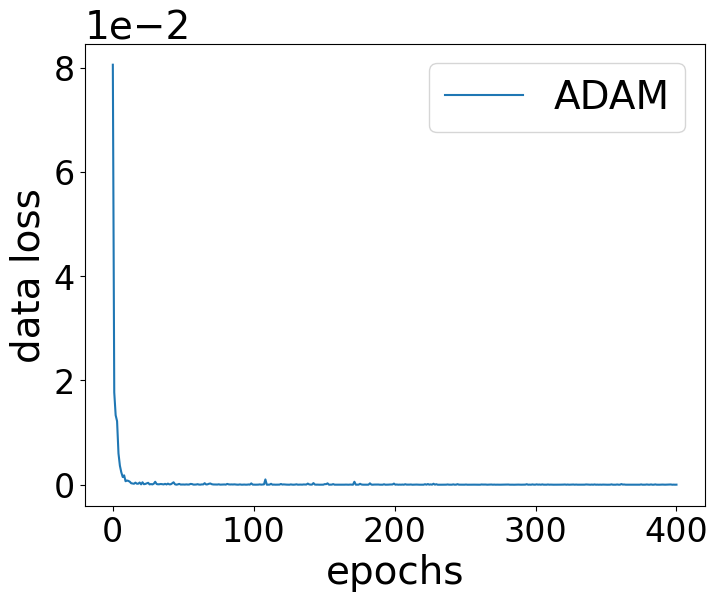}
        \includegraphics[width=.32\textwidth, height=3.cm]{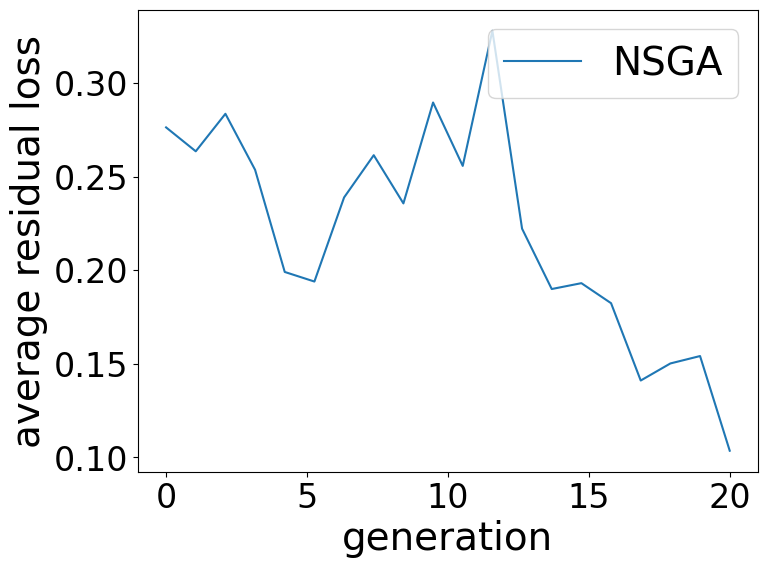}
        \includegraphics[width=.32\textwidth, height=3.cm]{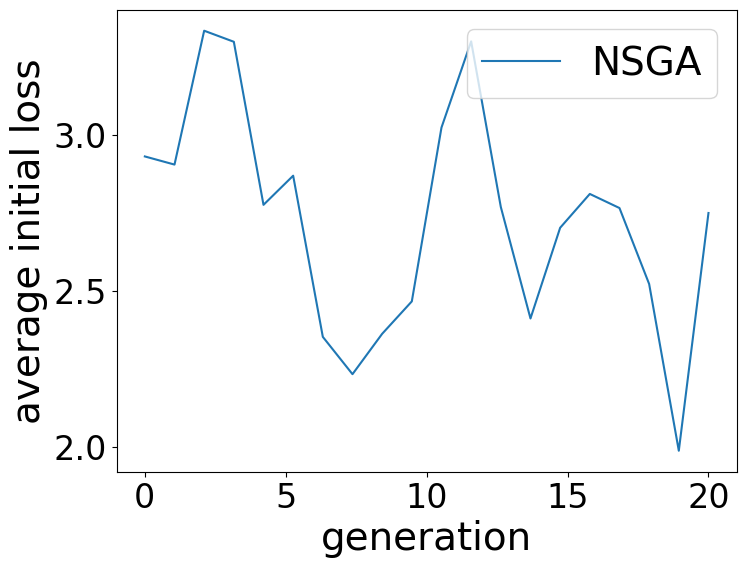}
        \includegraphics[width=.32\textwidth, height=3.cm]{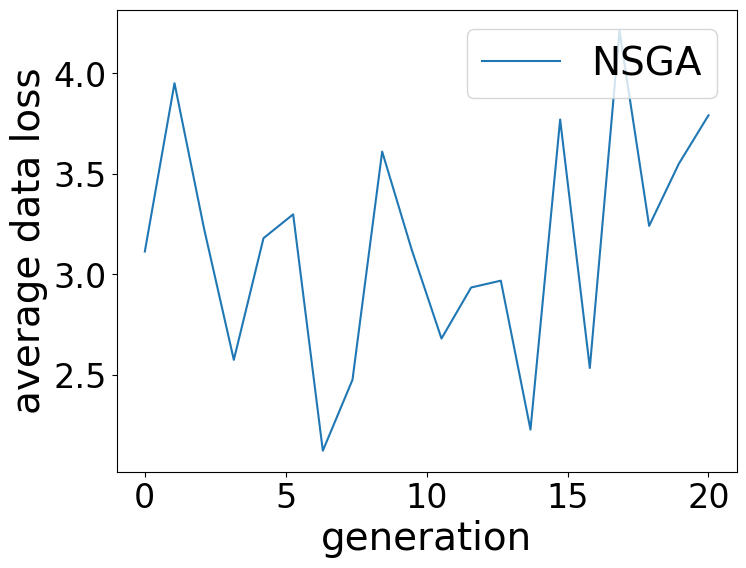}
        \includegraphics[width=.32\textwidth, height=3.cm]{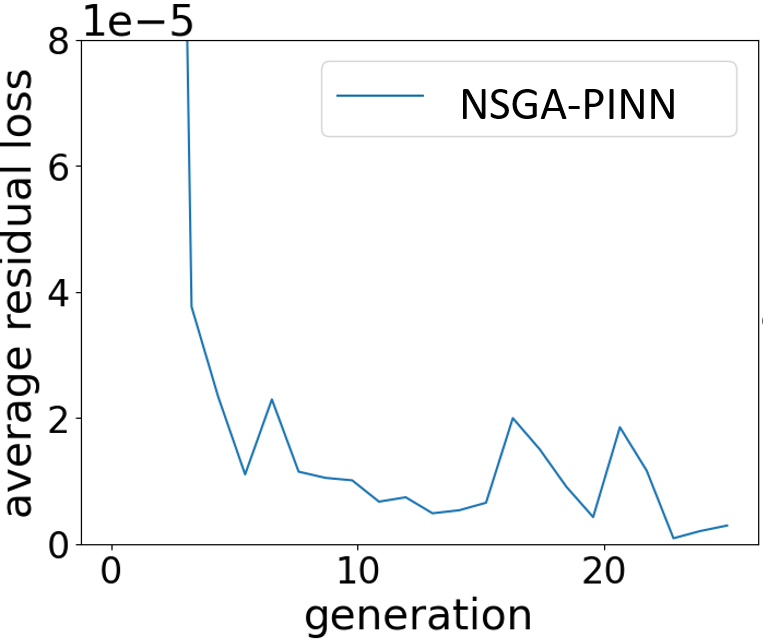}
        \includegraphics[width=.32\textwidth, height=3.cm]{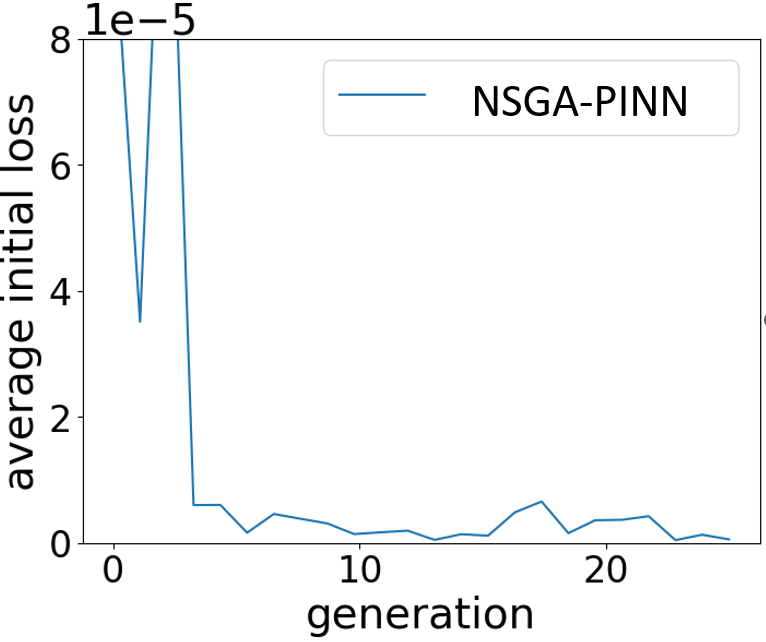}
        \includegraphics[width=.32\textwidth, height=3.cm]{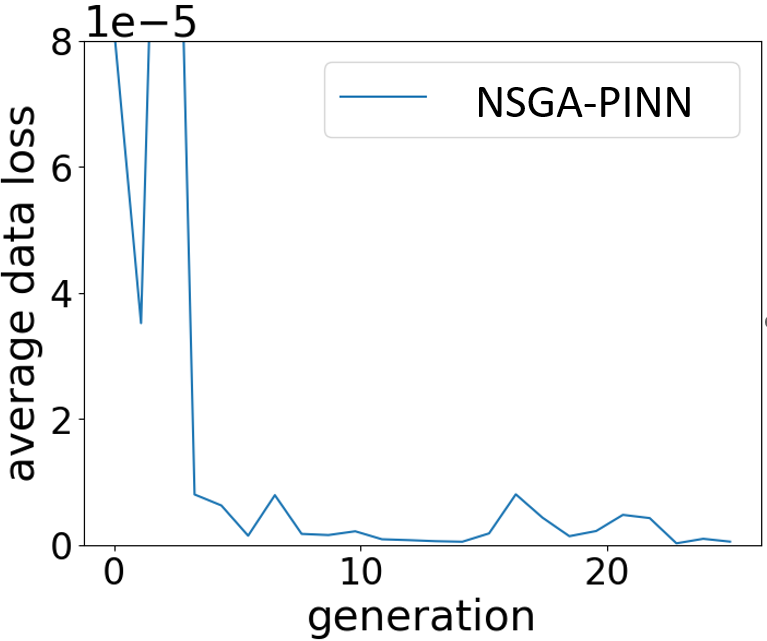}
     \hfill
     \caption{ Inverse pendulum problem: Group of figures from left to right columns show each loss value by using different training method.}
    \label{fig:inverse pendulum problem without data noise plot}
\end{figure}

In the course of our experiment, we tested various methods, which are illustrated in Figure~\ref{fig:inverse pendulum problem without data noise plot}. Based on our observations, we found that the Adam optimizer did not yield better results after 400 epochs, as the loss value remained around 1e-1. We also tried the NSGA-II algorithm for PINN, which introduced some diversity to prevent the algorithm from getting stuck at local minima, but the loss value was still around 4.0. Ultimately, we implemented our proposed NSGA-PINN algorithm to train PINN on this problem, resulting in a significant improvement with a loss value of 1e-5.

\begin{table} [h!]
\centering
\caption{Inverse pendulum problem: Each loss value from NN trained by using different training methods.  }
\begin{tabular} {l|cccc}
\hline 
\textbf{Methods} & \textbf{residual loss} & \textbf{Initial loss} & \textbf{data loss} & \textbf{total loss} \\
\hline
ADAM & 0.00013 & 3.24e-05 & 3.11e-05 &  1.935e-04\\
NSGA & 0.12    & 2.67     & 4.10 & 6.89 \\
NSGA-PINN & 8.96e-07 &  5.58e-07 & 5.22e-07 & 1.9760e-06\\
\hline
\end{tabular}
\label{table:predictive-errors}
\end{table}
To gain a clear understanding of the differences in loss values between optimization methods, we collected numerical loss values from our experiment. For the NSGA and NSGA-PINN methods, loss values were calculated as the average since they are obtained using ensemble methods through multiple runs. Our observations revealed that the total loss value of PINN trained with the traditional Adam optimizer decreased to 1.935e-04. However, by training with the NSGA-PINN method, the loss value decreased even further to 1.9760e-06, indicating improved satisfaction with the initial-condition constraints.

\begin{figure}[htb!]
     \centering
       
        \includegraphics[width=0.95\textwidth, height=3.5cm]{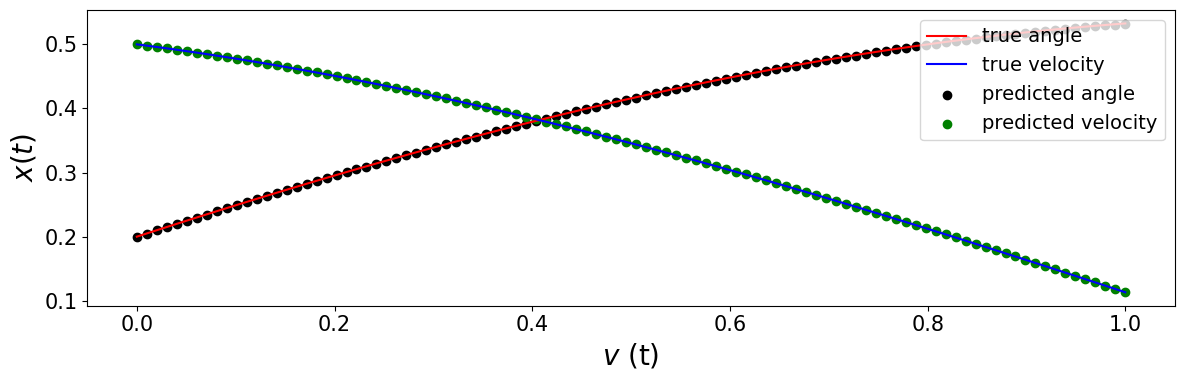}\\
        \includegraphics[width=0.95\textwidth, height=3.5cm]{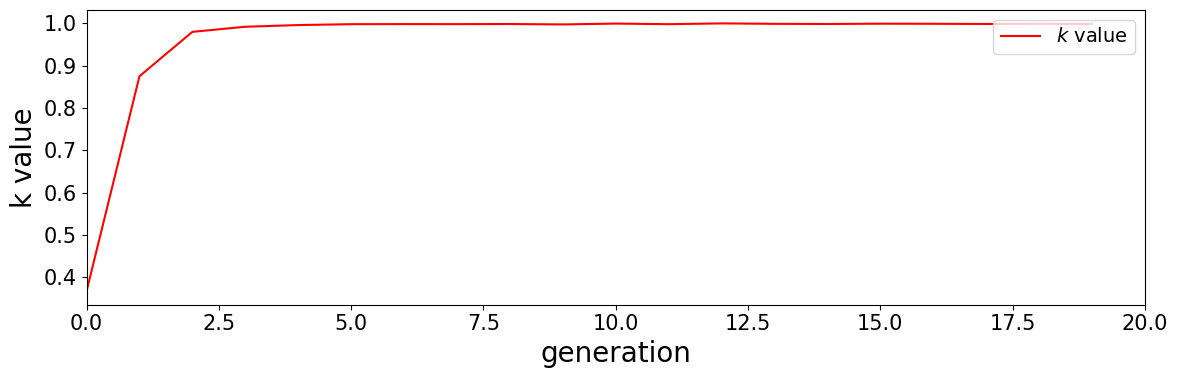}

     \hfill
         
     \caption{ Inverse pendulum problem: the top figure shows the comparison between the true value with the predicted value from PINN trained by NSGA-PINN method. The figure on the bottom shows the prediction of constant value k.  }
    \label{fig:inverse pendulum problem result without data noise }
\end{figure}

In Figure~\ref{fig:inverse pendulum problem result without data noise }, we compare the predicted angle and velocity state values to the true values to analyze the behavior of the proposed NSGA-PINN method. The top figure shows how accurately the predicted values match the true values, illustrating the successful performance of our algorithm. At the bottom of the figure, we observe the predicted value of the parameter $k$, which agrees with the true value of $k=1$. This result was obtained after running our NSGA-PINN algorithm for three generations.

\subsection{Inverse pendulum problem with noisy data}
In this section, we introduced Gaussian noise to the experimental data collected for the inverse problem. The noise was sampled from the Gaussian distribution:
\begin{align}
    P(x) = \frac{1}{{\sigma \sqrt {2\pi } }}e^{{{ - \left( {x - \mu } \right)^2 } \mathord{\left/ {\vphantom {{ - \left( {x - \mu } \right)^2 } {2\sigma ^2 }}} \right. \kern-\nulldelimiterspace} {2\sigma ^2 }}}
\end{align}
For this experiment, we chose to set the mean value ($\mu$) to 0 and the standard deviation of the noise ($\sigma$) to 0.1.

\begin{figure}
     \centering
        \includegraphics[width=.32\textwidth, height=3.cm]{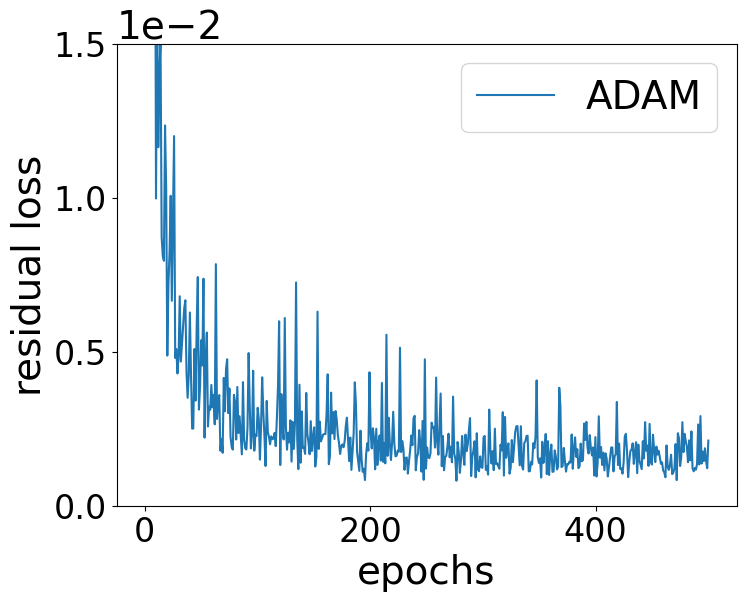} 
        \includegraphics[width=.32\textwidth, height=3.cm]{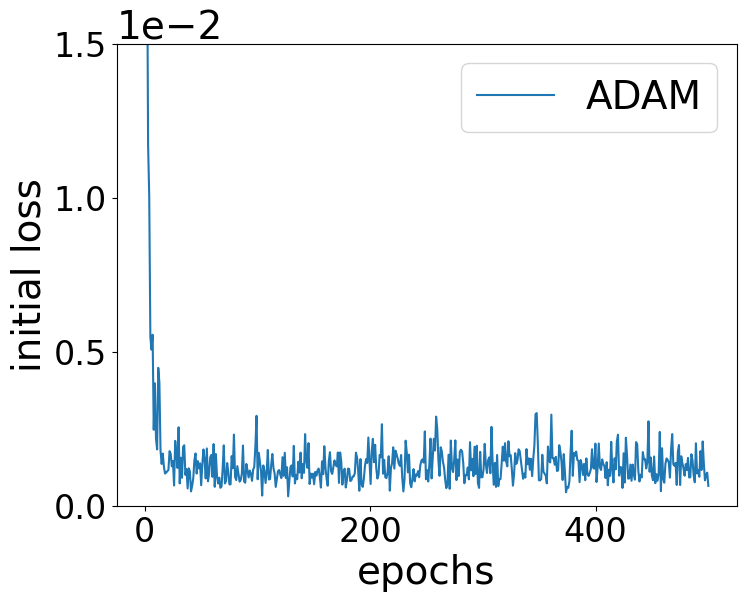}  
        \includegraphics[width=.32\textwidth, height=3.cm]{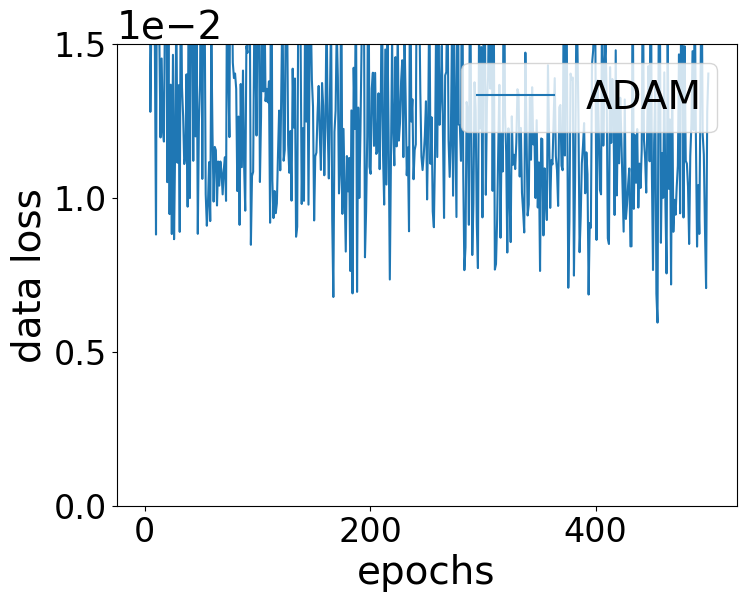}
       
        
        \includegraphics[width=.32\textwidth, height=3.cm]{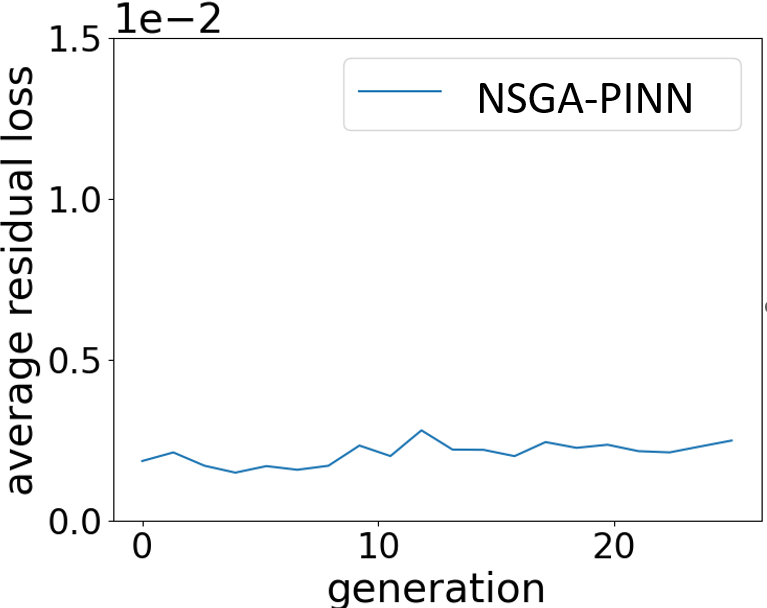}
        \includegraphics[width=.32\textwidth, height=3.cm]{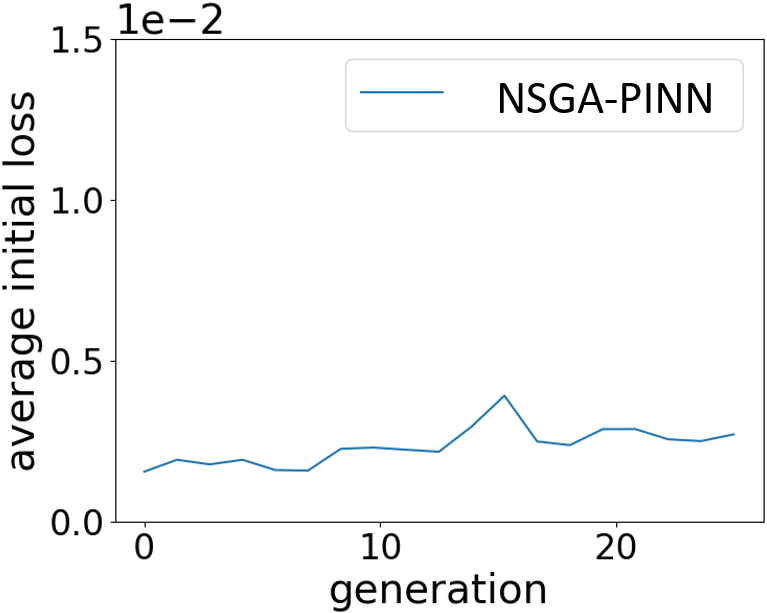}
        \includegraphics[width=.32\textwidth, height=3.cm]{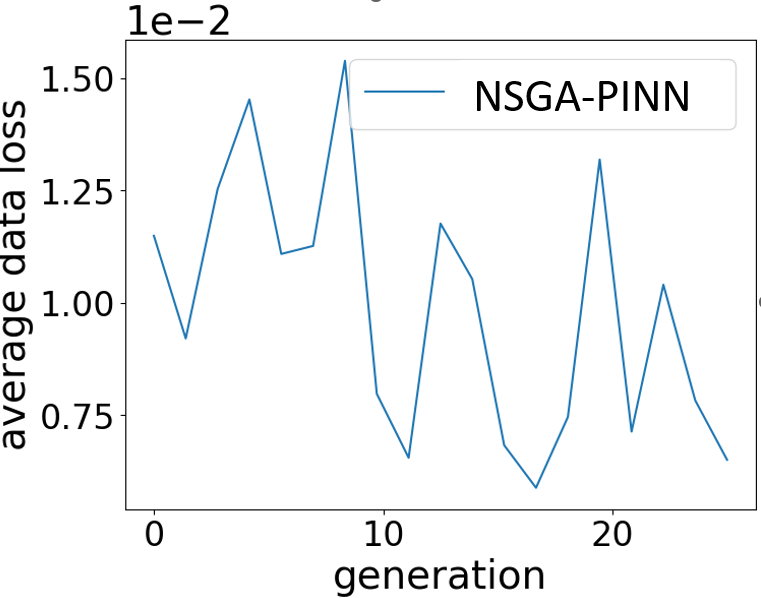}

     \hfill
         
     \caption{ Inverse pendulum problem with data noise: Group of figures from left to right columns show each loss value by using different training method with noisy data.  }
    \label{fig:inverse pendulum problem with data noise}
    \end{figure}
     
As depicted in Figure~\ref{fig:inverse pendulum problem with data noise}, we trained the PINN model using the Adam optimizer. However, we encountered an issue where the loss value failed to decrease after 400 epochs. This suggested that the optimizer had become stuck in a local minimum, which is a common problem associated with the Adam optimizer when presented with noise.

To address this issue, we implemented the proposed NSGA-PINN method, resulting in significant improvements. Specifically, by increasing the diversity of the NSGA population, we were able to escape the local minimum and converge to a better local optimum where the initial condition constraints were more effectively satisfied.
\begin{table} [h!]
\centering
\caption{Inverse pendulum problem with data noise: each loss value from NN trained by different training methods with noisy data.}
\begin{tabular} {l|cccc}
\hline 
\textbf{Methods} & \textbf{residual loss} & \textbf{Initial loss} & \textbf{data loss} & \textbf{total loss}\\
\hline
ADAM & 0.0028 & 0.0028 & 0.0114 & 0.017 \\

NSGA-PINN & 0.0015 &  0.0020 & 0.0089 & 0.0124\\

\hline
\end{tabular}
\label{table:Inverse pendulum problem with data noise: numerical results}
\end{table}

By examining Table~\ref{table:Inverse pendulum problem with data noise: numerical results}, we can see a clear numerical difference between the two methods. Specifically, the table shows that the PINN trained by the ADAM method has a total loss value of 0.017, while the PINN trained by the proposed NSGA-PINN method has a total loss value of 0.0124.

\begin{figure}[h!]
     \centering
        \includegraphics[width=1.0\textwidth, height=6cm]{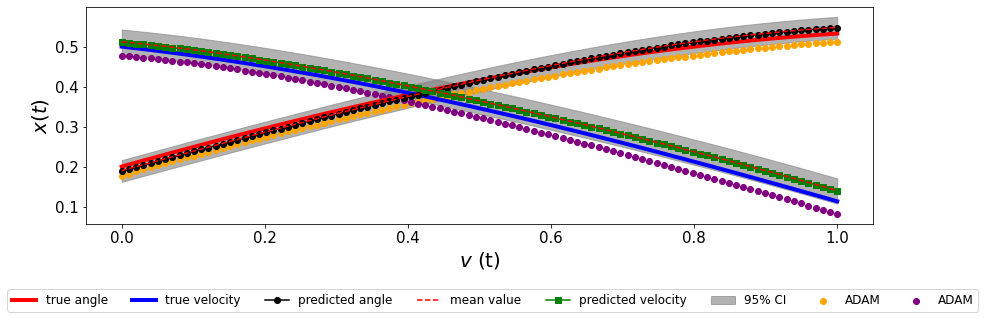} 
     \hfill 
     \caption{ Inverse pendulum problem with data noise : The figure shows the comparing of the result from PINN trained by NSGA-PINN method and ADAM method with noisy input data.}
    \label{fig:inverse pendulum problem result comparison with data noise}
\end{figure}

Finally, in Figure~\ref{fig:inverse pendulum problem result comparison with data noise}, we quantify uncertainty using an ensemble of predictions from our proposed method. This ensemble allows us to compute the 95\% confidence interval, providing a visual estimate of the uncertainty. To calculate the mean value, we averaged the predicted solutions from an ensemble of 100 PINNs trained by the NSGA-PINN algorithm. Our observations indicate that the mean is close to the solution, demonstrating the effectiveness of the proposed method. When comparing the predicted trajectory from the PINN trained with the NSGA-PINN algorithm to the one trained with the ADAM method, we found that the NSGA-PINN algorithm yields results closer to the real solution in this noisy scenario.

\subsection{Burgers equation}
This experiment uses the Burgers equation to study the effectiveness of the proposed NSGA-PINN algorithm on a PDE problem. The Burgers equation is defined as follows:
\begin{align}
   &\frac{du}{dt} + u \frac{du}{dx} = v\frac{d^2 u}{dx^2},\hspace{4mm}  x \in [-1,1], t \in [0,1]\\
   &u(0,x) = -\sin(\pi x) \nonumber \\
   &u(t,-1) = u(t,1) = 0 \nonumber
\end{align}
Here, $u$ is the PDE solution, $\Omega = [-1,1]$ is the spatial domain, and $v = 0.01/\pi$ is the diffusion coefficient.

The nonlinearity in the convection term causes the solution to become steep, due to the small value of the diffusion coefficient $v$. To address this problem, we utilized a neural network for PINN, which consisted of 8 hidden layers with 20 neurons each. The hyperbolic tangent activation function was used to activate the neurons in each layer. We sampled 100 data points on the boundaries and 10,000 collocation data points for PINN training.

For the proposed NSGA-PINN method, the original population size was set to 20 neural networks, and the algorithm ran for 20 generations. The loss function in the Burgers' equation can be defined as follows:
\begin{align*}
   \mathcal{L} = \mathcal{L}_{u} + \mathcal{L}_{b} + \mathcal{L}_{ics}.
\end{align*}
Here, the total loss value is the combination of the residual loss, the initial condition loss, and the boundary loss.

\begin{figure}[h!]
     \centering
        \includegraphics[width=.41\textwidth, height=4.cm]{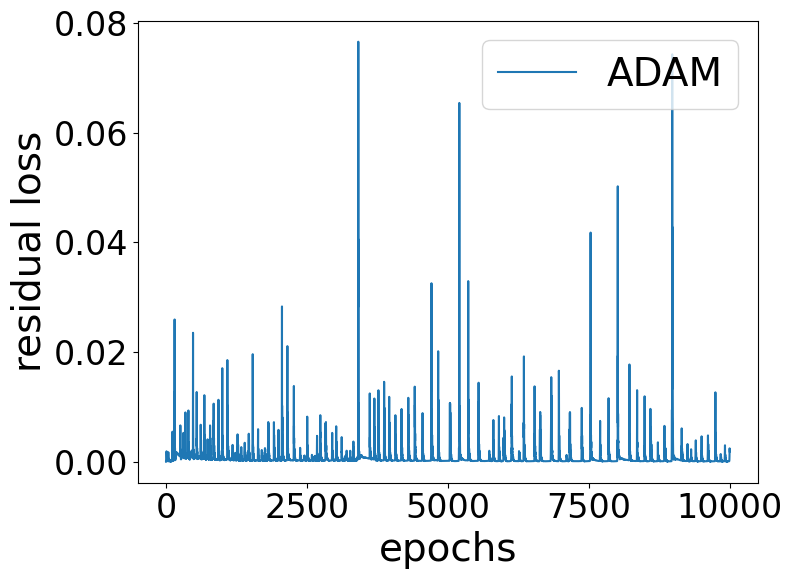} 
        \includegraphics[width=.41\textwidth, height=4.cm]{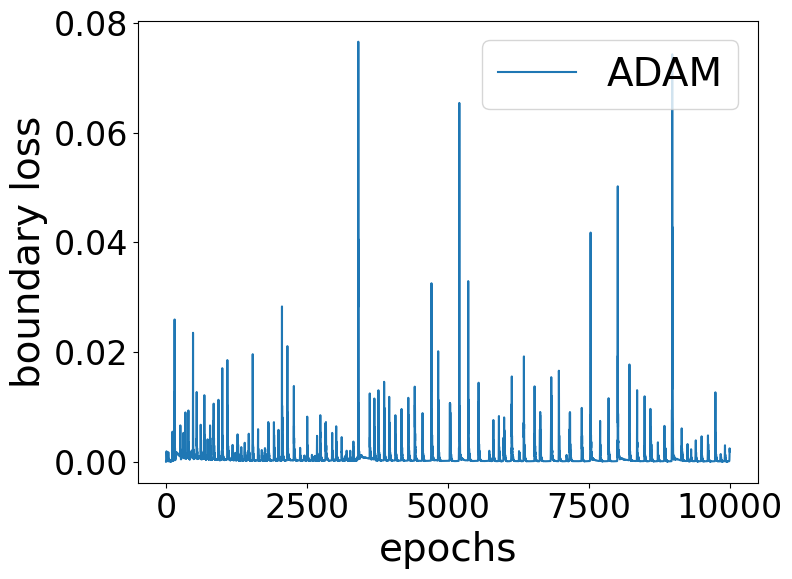}

        \includegraphics[width=.41\textwidth, height=4.cm]{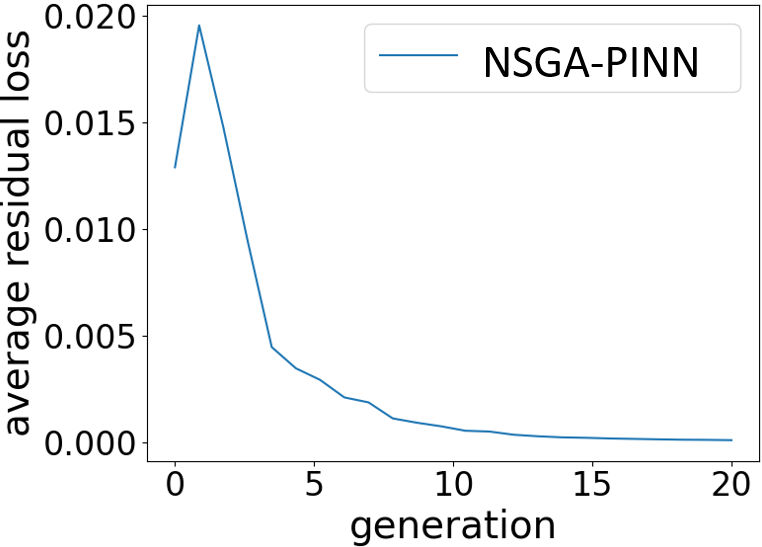}
        \includegraphics[width=.41\textwidth, height=4.cm]{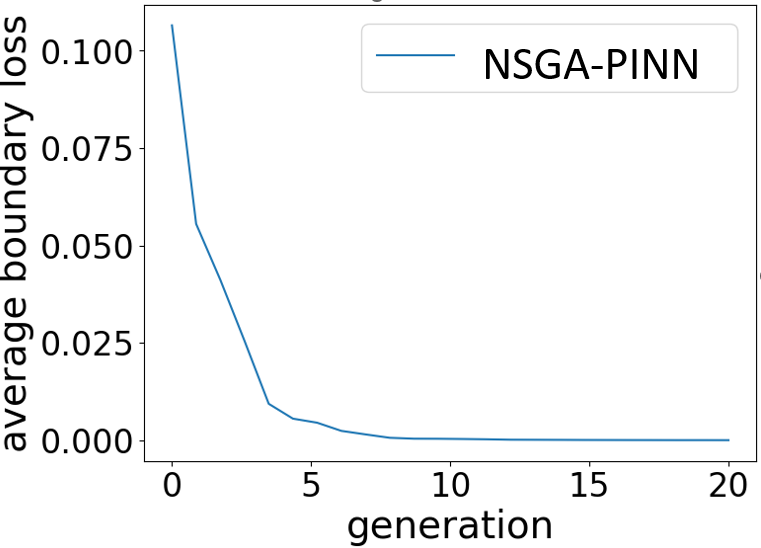}

     \hfill
         
     \caption{ Burgers equation: group of figures from left to right columns show each loss value by using different training method}
    \label{Burgers equation without noise: loss plot}
    \end{figure}

\begin{table} [h!]
\centering
\caption{Burgers equation: Comparison of the loss value from
NN trained by ADAM method and NSGA-PINN method for Burgers equation.}
\begin{tabular} {l|ccc}
\hline 
\textbf{Methods} & \textbf{residual loss} &\textbf{boundary loss} & \textbf{total loss}\\
\hline
ADAM & 0.0002 & 9.4213e-05 & 0.0003 \\

NSGA-PINN & 0.0001& 7.0643e-05 & 0.0002\\

\hline
\end{tabular}
\label{table:Burgers equation loss value without data loss}
\end{table}

\begin{figure}[h!]
     \centering
         \includegraphics[width=12cm, height=6cm]{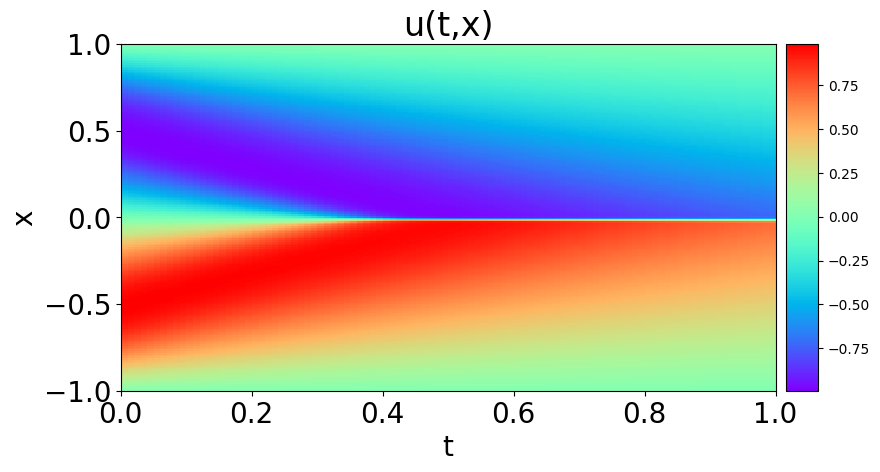}\\
          \includegraphics[width=.32\textwidth, height=3.2cm]{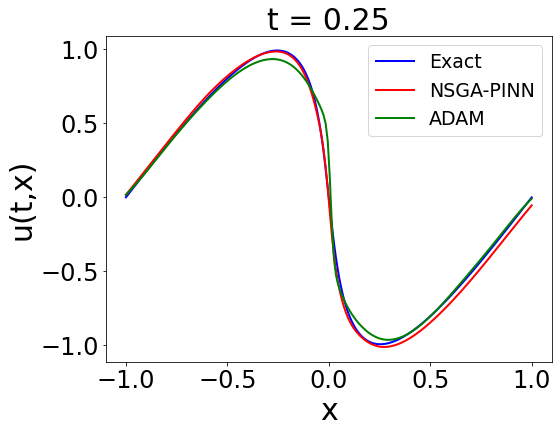}
          \includegraphics[width=.32\textwidth, height=3.2cm]{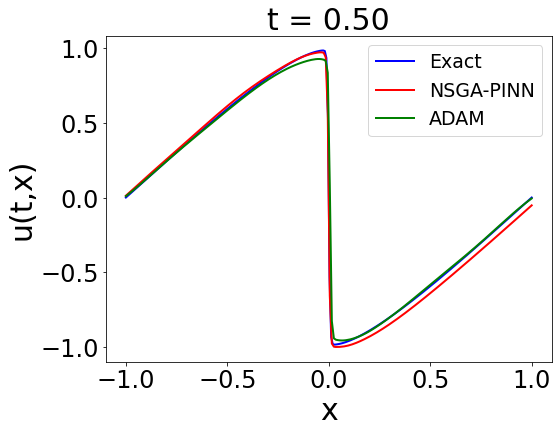}
          \includegraphics[width=.32\textwidth, height=3.2cm]{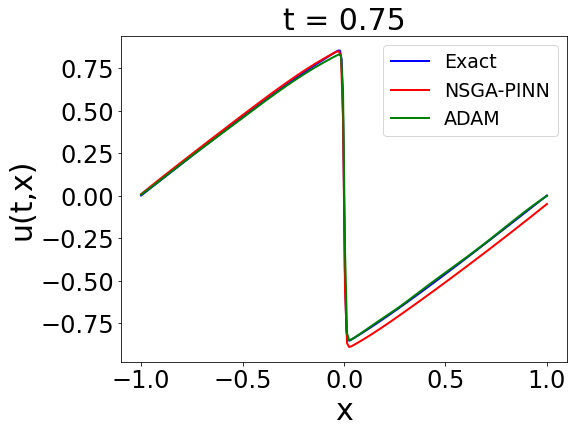}
     \hfill
         
     \caption{Burgers equation : the top panel shows the contour plots of solution of the Burgers equation. The lower figure shows the comparison of the exact value with the predicted value at different time point.}
     \label{Burgers equation result without data noise}
\end{figure}

We can observe the effectiveness of the proposed NSGA-PINN algorithm by examining the loss values depicted in Figure~\ref{Burgers equation result without data noise} and Table~\ref{table:Burgers equation loss value without data loss}. In particular, Table ~\ref{table:Burgers equation loss value without data loss} compares the loss values of PINNs trained by the NSGA-PINN algorithm and the traditional ADAM method. The results show that the ADAM and NSGA-PINN algorithms produce almost identical loss values.

Finally, Figure~\ref{Burgers equation result without data noise} displays contour plots of the solution to Burgers' equation. The top figure shows the result predicted using the proposed NSGA-PINN algorithm. The bottom row compares the exact value with the values from the proposed algorithm and the ADAM method at t = 0.25, 0.50, and 0.75. Based on this comparison, both the NSGA-PINN algorithm and the ADAM method predict values that are close to the true values.

\subsection{Burgers equation with noisy data}
\label{Burgers equation (noisy data)}
In this experiment, we evaluate the effectiveness of the NSGA-PINN algorithm when applied to noisy data and the Burgers equation. We compare the results obtained from the proposed algorithm with those obtained using the ADAM optimization algorithm. To simulate a noisy scenario, Gaussian noise was added to the experimental/input data. We sampled the noise from a Gaussian distribution with a mean value ($\mu$) of 0.0 and a standard deviation ($\sigma$) of 0.2. 

\begin{figure}[h!]
     \centering
        \includegraphics[width=.41\textwidth, height=4.cm]{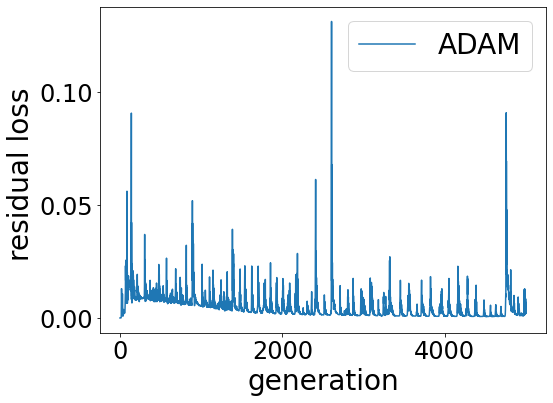} 
        \includegraphics[width=.41\textwidth, height=4.cm]{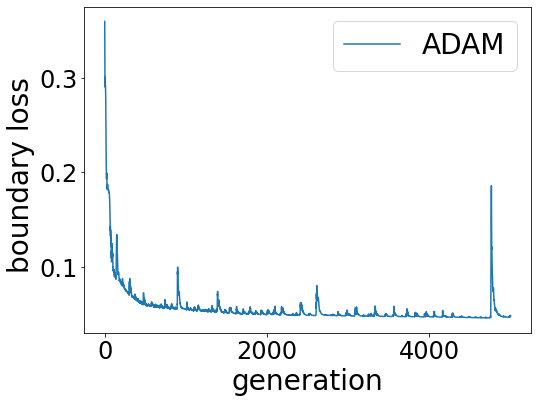}

        \includegraphics[width=.41\textwidth, height=4.cm]{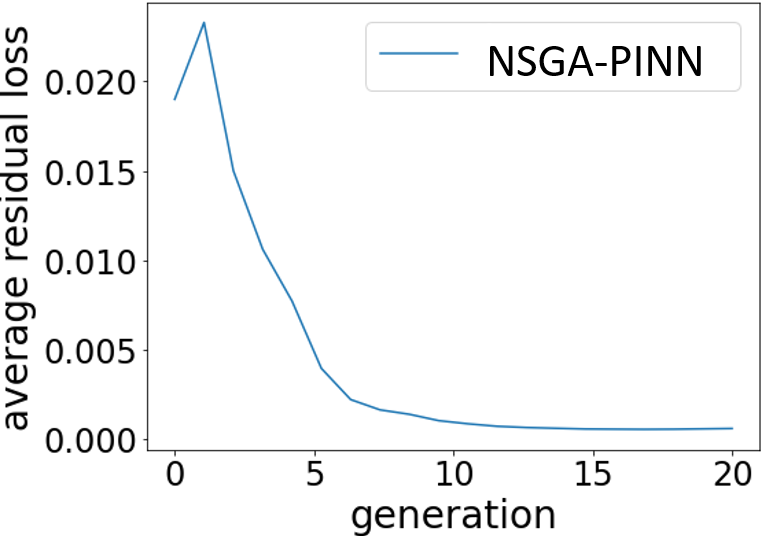}
        \includegraphics[width=.41\textwidth, height=4.cm]{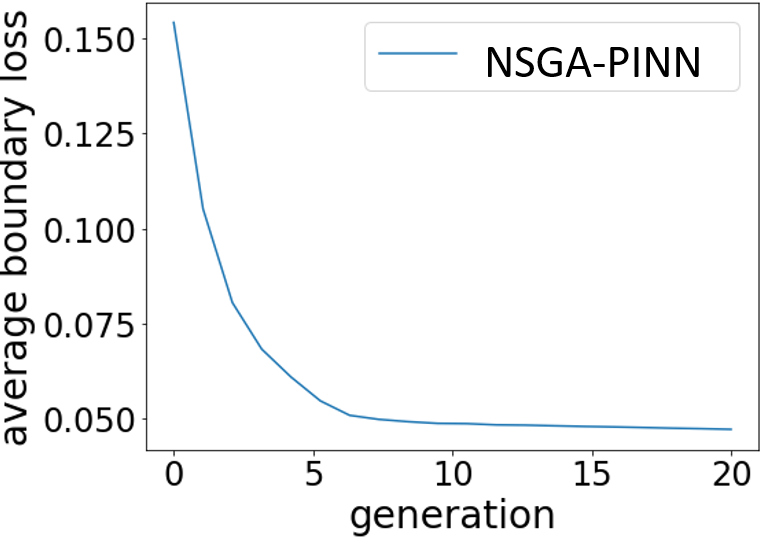}

     \hfill
         
     \caption{ Burgers equation with data noise: the left column shows the residual loss. The right column shows the boundary loss.
    }
    \label{Burgers equation loss value plot with data noise}
\end{figure}

\begin{table} [h!]
\centering
\caption{Burgers equation with data noise: Comparison of the loss value from NN trained by ADAM method and NSGA-PINN method for Burgers equation with noisy data.}
\begin{tabular} {l|ccc}
\hline 
\textbf{Methods} & \textbf{residual loss} &\textbf{boundary loss} & \textbf{total loss}\\
\hline
ADAM & 0.0045 & 0.0481 & 0.0526 \\

NSGA-PINN & 0.0006& 0.0473 & 0.0479\\

\hline
\end{tabular}
\label{table:Burgers equation loss vlaue with data noise}
\end{table}

We analyzed the effectiveness of the proposed NSGA-PINN method with noisy data. Specifically, Figure~\ref{Burgers equation loss value plot with data noise} and Table~\ref{table:Burgers equation loss vlaue with data noise} illustrate the corresponding loss values. It is worth noting that, while the PINN trained with ADAM no longer improves after 5000 epochs and reaches a final loss value of 0.0526, training the PINN with the proposed algorithm for 20 generations results in a reduced total loss of 0.0479.

\begin{figure}
     \centering
         \includegraphics[width=12cm, height=6cm]{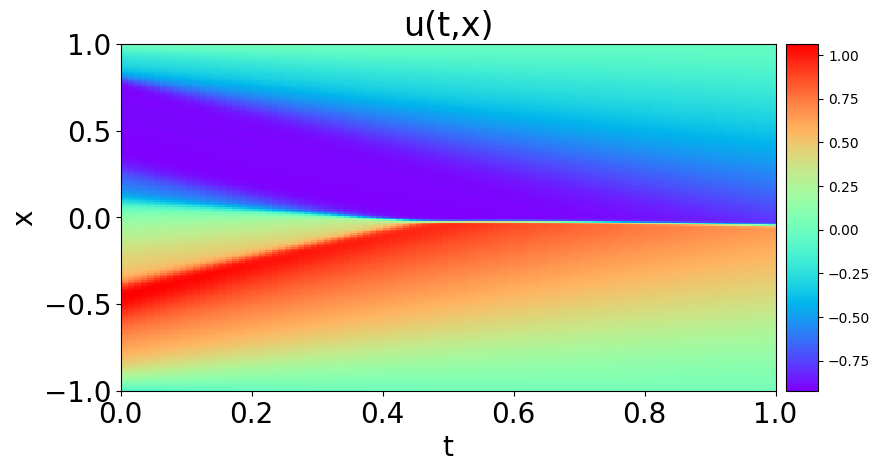}\\
          \includegraphics[width=.32\textwidth, height=3.2cm]{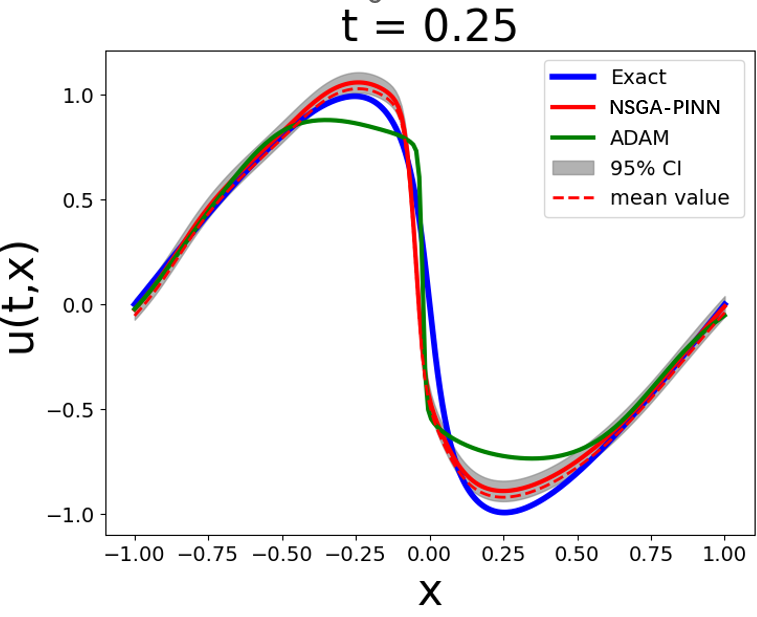}
          \includegraphics[width=.32\textwidth, height=3.2cm]{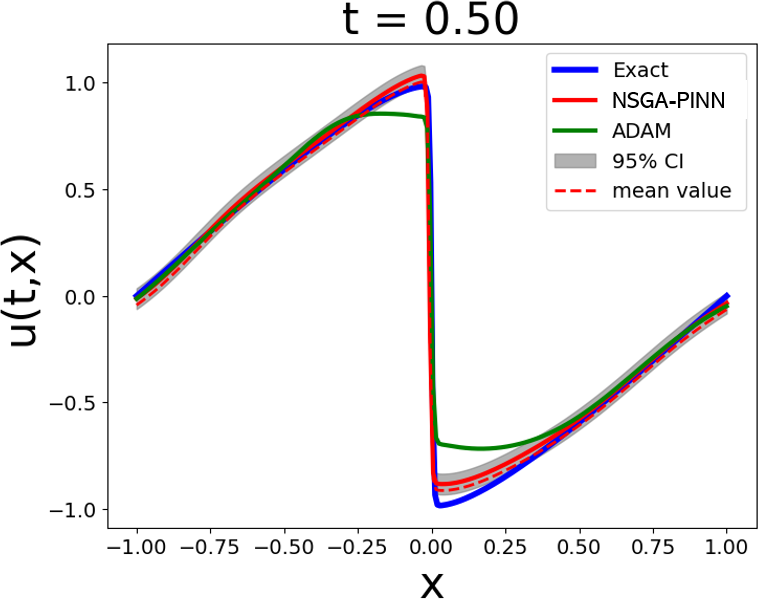}
          \includegraphics[width=.32\textwidth, height=3.2cm]{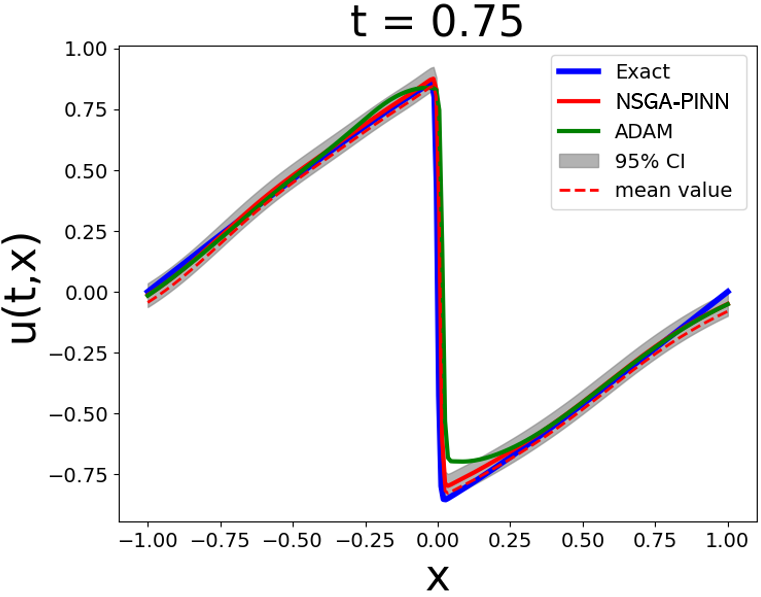}
     \hfill
         
     \caption{Burgers equation with data noise: the top panel shows the contour plots of solution of the Burgers equation. The lower figure shows the comparison of the exact value with the predicted value at different time point.}
     \label{Burgers equation with data noise result}
\end{figure}

Finally, Figure~\ref{Burgers equation with data noise result} shows the results of the PINN trained by the NSGA-PINN method with noisy data. The top figure shows a smooth transition over space and time. The lower figures compare the true value with the predicted value for the PINN trained by the proposed method and the traditional ADAM optimization algorithm. The results demonstrate that the prediction from a PINN trained by NSGA-PINN approaches the true value of the PDE solution more closely.

\subsection{Test survival rate}
 In this final experiment, we conducted further tests to verify the feasibility of our algorithm. Specifically, we calculated the survival rate between each generation to determine if the algorithm was learning and using the learned results as a starting point for the next generation.

The experiment consisted of the following steps: First, we ran the total NSGA-PINN method 50 times. Then, for each run, we calculated the survival rate between each generation using the following formula:
 \begin{align}
   &S = Q_i / P_i.
\end{align}
\label{survival rate}
Here, $Q_i$ represents the number of offspring from the previous generation, and $P_i$ represents the number of parent population in the current generation. Finally, to obtain a relatively robust data that represents the trend of survival rate, we calculate the average value of survival rate between each generation as the algorithm progresses.

\begin{figure}[h!]
     \centering
        \includegraphics[width=.8\textwidth, height=10.cm]{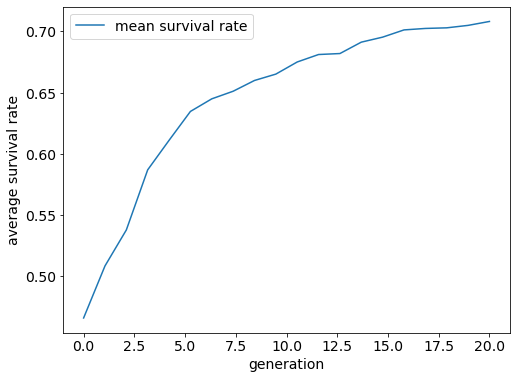}
     \hfill
         
     \caption{ survival rate between each generation}
    \label{fig:test survival rate}
\end{figure}
Figure~\ref{fig:test survival rate} shows that the survival rate increases as the algorithm progresses. The survival rate of the first two generations is approximately 50\%, but by the end of the algorithm, it improves to 71\%. This indicates that our algorithm is progressively learning as subsequent generations are generated, which significantly enhances PINN training.
\section{Discussion} \label{sec:discussion}
The experimental results in the previous section showed promising outcomes for training PINNs using the proposed NSGA-PINN method. As described in Section ~\ref{sec:proposed-method}, when solving the inverse problem using the traditional Adam optimizer, the algorithm became trapped in a local optimum after running for 400 epochs. However, by using the NSGA-PINN method, the loss value continued to decrease, and the predicted solution was very close to the true value. Additionally, when dealing with noisy data, the traditional Adam optimizer had difficulty learning quickly and making accurate predictions. On the other hand, the proposed NSGA-PINN algorithm learned efficiently and converged to a better local optimum for generalization purposes.

However, the main drawback of the proposed method is that it requires an ensemble of neural networks (NNs) during training. Consequently, the proposed NSGA-PINN incurs a larger computational cost than traditional stochastic gradient descent methods. Therefore, reducing the computational cost of NSGA-PINN is a goal for our future work. For instance, some of the training computational cost could be mitigated by using parallelization. Additionally, we will attempt to derive effective methods for finding the best trade-off between NSGA and Adam.


More specifically, in our future work, we will focus on balancing the parent population ($N$), max generation number ($\alpha$), and number of epochs used in the Adam optimizer. These values are manually initialized in the proposed method. The parent population determines the diversity in the algorithm, and we ideally want high diversity. The max generation number determines the total learning time. Increasing this time allows the algorithm to continue learning from previous generations, but it may lead to overfitting if the number is too large. Note that there is a trade-off between the max generation number and the epoch number used in the Adam optimizer. A higher generation number allows the NSGA algorithm to perform better, helping the Adam optimizer escape local optima, but this comes at a higher computational cost. Meanwhile, increasing the number of epochs used in the Adam optimizer helps the model decrease the loss value quickly, but it reduces the search space and may lead to the algorithm becoming trapped in local minima.

\section{Conclusion} \label{sec:conclusion}
In this paper, we proposed a novel multi-objective optimization method called NSGA-PINN for training physics-informed neural networks. Our approach involves using the Non-dominated Sorting Genetic Algorithm (NSGA) to handle each component of the training loss in PINN. This allows us to achieve better results in terms of inverse problems, noisy data, and satisfying constraints. We demonstrated the effectiveness of NSGA-PINN by applying it to several ordinary and partial differential equation inverse problems. Our results show that the proposed framework can handle challenging noisy scenarios.

\bibliographystyle{unsrt}  

\end{document}